\documentclass[10pt,twocolumn,letterpaper]{article}

\usepackage{wacv}              

\definecolor{wacvblue}{rgb}{0.21,0.49,0.74}
\usepackage[pagebackref,breaklinks,colorlinks,allcolors=wacvblue]{hyperref}
\usepackage{multirow}
\def\wacvPaperID{1833} 
\def\confName{WACV}
\def\confYear{2026}

\title{PredMapNet: Future and Historical Reasoning for Consistent Online HD Vectorized Map Construction}
\author{
Bo Lang$^{1}$ \quad
Nirav Savaliya$^{2}$ \quad
Zhihao Zheng$^{1}$ \quad
Jinglun Feng$^{2}$ \quad
Zheng-Hang Yeh$^{2}$ \quad
Mooi Choo Chuah$^{1}$ \\
$^{1}$Lehigh University \quad
$^{2}$Honda Research Institute USA \\
{\tt\small \{bol221, zhzc21, mcc7\}@lehigh.edu} \\
{\tt\small \{nsavaliya, jinglun\_feng, zheng-hang\_yeh\}@honda-ri.com}
}

\begin{document}
\maketitle
\begin{abstract}

High-definition (HD) maps are crucial to autonomous driving, providing structured representations of road elements to support navigation and planning. However, existing query-based methods often employ random query initialization and depend on implicit temporal modeling, which lead to temporal inconsistencies and instabilities during the construction of a global map. To overcome these challenges, we introduce a novel end-to-end framework for consistent online HD vectorized map construction, which jointly performs map instance tracking and short-term prediction. First, we propose a Semantic-Aware Query Generator that initializes queries with spatially aligned semantic masks to capture scene-level context globally. Next, we design a History Rasterized Map Memory to store fine-grained instance-level maps for each tracked instance, enabling explicit historical priors. A History-Map Guidance Module then integrates rasterized map information into track queries, improving temporal continuity. Finally, we propose a Short-Term Future Guidance module to forecast the immediate motion of map instances based on the stored history trajectories. These predicted future locations serve as hints for tracked instances to further avoid implausible predictions and keep temporal consistency. Extensive experiments on the nuScenes and Argoverse2 datasets demonstrate that our proposed method outperforms state-of-the-art (SOTA) methods with good efficiency. All source code will be publicly released. More information can be found on our project page: \url{https://astronirav.github.io/predmapnet}
\end{abstract}

\section{Introduction}
\label{sec:intro}

High-definition (HD) maps are critical for map-based autonomous driving, providing rich semantic and geometric context of the environment. Traditionally, HD maps are constructed through point cloud-based SLAM systems—using either Camera-LiDAR fusion \cite{LVI-SAM} or LiDAR-only approaches \cite{LOAM, LeGOLOAM, LIOSAM, GOLO}—followed by manual annotation of semantic elements such as lane boundaries, dividers, crosswalks, and directionality. While highly accurate, this pipeline is labor-intensive, expensive, and difficult to scale. Recently, deep learning-based methods have emerged as a scalable alternative, enabling online vectorized HD map construction from camera-only or camera-LiDAR fusion inputs \cite{HDMapNet, BEVFusion}. These approaches offer a cost-effective solution for building large-scale city maps and, due to their online nature, support deployment in unseen environments and facilitate dynamic map change detection \cite{MapChange}.

\begin{figure}[t]
\centering
\includegraphics[width=0.45\textwidth]{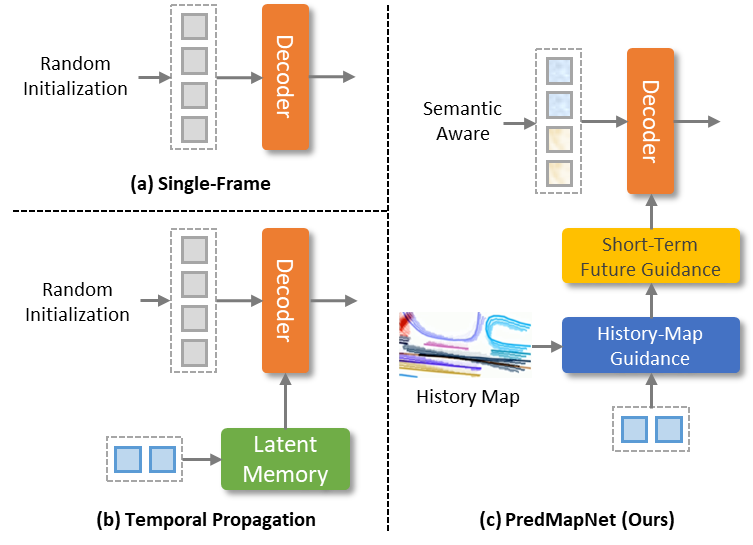}
\caption{Comparison of several HD map construction methods: (a) Single-Frame, (b) Temporal Propagation, (c) PredMapNet (Ours) utilizes semantic-aware information and enables both historical \& future reasoning for better query decoding.}
\label{shortD}
\vspace{-1em}
\end{figure} 

Early methods approached vectorized HD map construction either as a per-frame BEV rasterization task \cite{HDMapNet, LiftSplatShoot} or as a DETR-style point set prediction and aggregation problem \cite{VectorMapNet, MapTR, MapTRv2}. BEV rasterization methods typically require post-processing to extract vectorized map elements. Other trends leveraged the DETR-based detection paradigm \cite{DETR} to  directly predict single-frame vectorized map components by decoding learnable queries from the BEV features in parallel, as shown in Fig. \ref{shortD} (a). However, due to the elongated and structured nature of road map elements, deformable DETR \cite{DeformableDETR} based models often struggle to  capture the semantic and geometric information of map instances in complex scenes. To address this, later works \cite{BeMapNet, PivotNet, HIMap, GeMap} incorporated geometric priors, leveraging the fact that map elements follow well-defined shapes. While this improved structural consistency, these methods still suffered from limited performance, primarily because they operated on individual frames without propagating temporal information from previous predictions. Recent methods such as \cite{StreamMapNet} and \cite{MapTracker}, as shown in Fig. \ref{shortD} (b), propose to maintain a separate memory modules to ensure temporal consistency in predictions. \cite{StreamMapNet} proposes to use streaming fusion to maintain a single latent memory whereas MapTracker \cite{MapTracker} uses separate BEV Raster and Vector latent memories. 

In this study, we propose an enhanced vectorized HD map construction framework, referred to as PredMapNet, by introducing three key ideas. As illustrated in Fig. \ref{shortD} (c), First, we adopt a Semantic-Aware Query Generation (SAQG) strategy inspired by Mask2Former \cite{Mask2Former}, which utilizes class-agnostic BEV segmentation masks to guide the refinement of queries. Unlike random initialization, this approach leverages global semantic context to produce context-aligned queries, improving query quality and training convergence. Second, we incorporate historical rasterized map information via a History-Map Guidance (HMG) module, enabling smoother and more continuous detection and better utilization of temporal priors. Finally, we are the first to introduce short-term future reasoning into online HD map construction. We design a Short-Term Future Guidance (STFG) module that explicitly predicts the near-future positions of map instances. By injecting motion priors, STFG improves the temporal stability of tracked queries and ensures more coherent instance predictions across consecutive frames.



In summary, our main contributions are as follows:
\begin{itemize}
    \item We propose PredMapNet, a novel end-to-end framework for consistent online HD map construction. Our model relies on a Semantic-Aware Query Generator to capture semantic information for map instances, enhancing spatial alignment and semantic awareness.
    
    \item We are the first to incorporate short-term future reasoning into this task. To this end, we introduce a unified framework that incorporates both historical and future guidance. The History-Map Guidance Module integrates fine-grained historical priors from history maps, while the Short-Term Future Guidance Module forecasts near-future positions of map instances to inject motion-aware priors. Together, these modules enhance temporal coherence and enable robust query propagation.

    \item PredMapNet achieves new state-of-the-art results on two representative benchmarks of online vectorized HD map construction, validating the effectiveness of the proposed modules.
\end{itemize}
\section{Related Works}
\label{sec:formatting}
\subsection{Online Vectorized HD Map Construction}

The task of online vectorized HD map construction has received increasing attention in recent years. HDMapNet \cite{HDMapNet} pioneered this line of work by lifting surround-view image features into the BEV space and extracting a rasterized BEV representation, which is later post-processed to obtain vectorized map elements. In contrast, VectorMapNet \cite{VectorMapNet} and MapTR \cite{MapTR} formulate the task as a DETR-style \cite{DETR} point set prediction problem. VectorMapNet first detects map elements and then autoregressively generates polylines for each instance. MapTR introduces a permutation-invariant formulation by designing a hierarchical query embedding scheme, performing bipartite matching at both instance and point levels to predict all vertices simultaneously. MapTRv2 \cite{MapTRv2} builds upon MapTR by decoupling self-attention operations to reduce memory usage and computational complexity.

More recent geometry-aware approaches \cite{PivotNet, BeMapNet, GeMap, HIMap} improve upon prior methods by incorporating inductive biases from map element geometries. BeMapNet \cite{BeMapNet} represents elements as piecewise Bézier curves, while GeMap \cite{GeMap} learns geometry-aware features that are invariant to translation and rotation. PivotNet \cite{PivotNet} follows the set prediction paradigm and models each map element using a dynamic set of pivotal points. HIMap \cite{HIMap} introduces a novel hybrid representation, enabling joint learning from both rasterized and vectorized formats through a proposed point-element interaction mechanism.

\begin{figure*}[t]
\centering
\includegraphics[width=\textwidth]{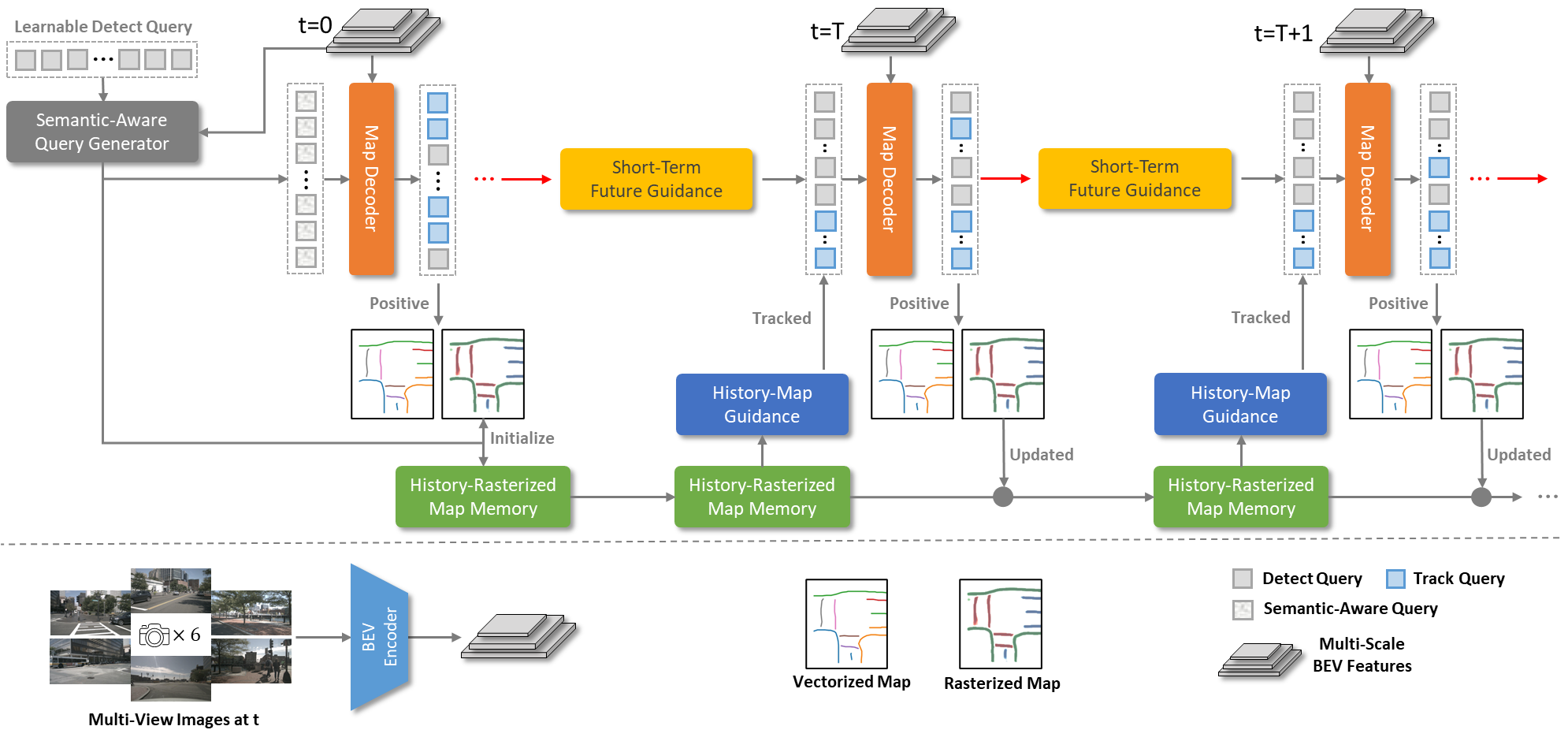}
\caption{The architecture pipeline of PredMapNet. At each frame, multi-view images are processed by a BEV encoder to extract perception features. The Semantic-Aware Query Generator \ref{SAQG} produces semantic-aware detection queries and rasterized map from BEV features. A History Rasterized Map Memory \ref{memory} is maintained to store instance-level segmentation masks over time. The History-Map Guidance Module \ref{HMG} refines track queries with historical geometric priors from memory. Simultaneously, the Short-Term Future Guidance Module \ref{STFG} predicts future polylines from historical trajectories and fuses them into track queries to guide query initialization in the next frame. Together, these modules enable temporally consistent and robust map instance construction across frames.}
\label{framework}
\vspace{-1em}
\end{figure*}
\subsection{Temporal Modeling in Perspective \& BEV Perception}

In visual object tracking, the query tracking paradigm has gained significant popularity in recent years, with methods such as TrackFormer \cite{TrackFormer}, TransTrack \cite{TransTrack}, and MOTR \cite{MOTR, MOTRv2}. These methods have been further extended in MeMOT \cite{MeMOT} and MeMOTR \cite{MeMOTR} by introducing memory modules to ensure long-range temporal consistency.

In BEV perception, there are two fundamental strategies for memory fusion: stacking and streaming. Stacking-based approaches, such as BEVDet4D \cite{BEVDet4D} and BEVFormer v2 \cite{BEVFormerv2}, process multiple historical frames in a single pass. While effective, this design incurs substantial memory and computation costs that scale with the number of input frames, limiting scalability and restricting temporal modeling to short-term horizons. In contrast, the streaming fusion strategy \cite{VideoBEV, StreamPETR, Sparse4Dv2, lang2024bev} processes frames sequentially, propagating memory features from previous frames. This enables longer temporal associations with reduced memory usage and lower latency.

StreamMapNet \cite{StreamMapNet} extends the single-frame vectorized HD map construction paradigm to a temporally consistent mapping framework. It leverages a streaming fusion strategy inspired by \cite{VideoBEV, StreamPETR, Sparse4Dv2} and introduces a novel Multi-Point Attention mechanism to handle irregular and elongated map elements. SQD-MapNet \cite{SQDMapNet} builds upon StreamMapNet by incorporating temporal curve denoising, inspired by DN-DETR \cite{DNDETR}. MapTracker \cite{MapTracker} enhances memory modules by stacking two types of memory buffers—rasterized and vectorized BEV memories—and adopts a query propagation paradigm for tracking, following the lineage of \cite{TrackFormer, TransTrack, MOTR, MOTRv2}. PrevPredMap \cite{PrevPredMap} introduces a lightweight approach to improve temporal consistency by reusing previous predictions as memory for future frames.

 \section{Methodology}
\subsection{Architecture Overview}
Our proposed framework aims to construct both globally consistent rasterized and vectorized maps, following a query-based tracking and prediction paradigm, as illustrated in Fig. \ref{framework}. At each frame, multi-view images are processed by the BEV encoder to extract multi-scale bird’s-eye-view (BEV) features $F_{bev}$. The Semantic-Aware Query Generator extracts semantic-aware queries from BEV Features using the Mask-aware attention and generates rasterized map (BEV segmentation masks) through multi-layer decoding. Then, the perception features together with the semantic-aware detection queries $Q \in \mathbb{R}^{N_q\times C}$ interact through the map decoder to produce the coordinates $P \in \mathbb{R}^{N_q\times N_p \times 2}$ , categories $C \in \mathbb{R}^{N_q \times 3}$ (pedestrian crossing, road boundary, lane divider) and scores $S \in \mathbb{R}^{N_q}$. Here, $N_q$ is the number of detected queries, $N_p$ is the number of points per map instance. Subsequently, the predicted detection queries are filtered using a confidence threshold $\tau_{d}$ while track queries associated with the corresponding detection queries from the previous frame are further filtered using a tracking threshold $\tau_{t}$. These positive $N_{track}$ queries $Q_{track}$ are carried forward to the next frame to maintain instance continuity.  

Inspired by HRMapNet \cite{zhang2024enhancing}, we maintain a History Rasterized Map Memory to support temporal consistency, which explicitly stores the predicted instance-level rasterized map from previous frames. For each tracked instance, this memory preserves spatial information over time, enabling structured temporal propagation. To effectively utilize this memory, the History-Map Guidance Module samples region-specific features from the global BEV features.
These sampled features provide strong geometric priors for enhancing the accuracy and continuity of track queries in the current frame. 

In addition, the Short-Term Future Guidance Module takes the history trajectories of the positive $N_{track}$ queries $Q_{track}$ as input and predicts short-term future positions of each query. In the next new frame, these predicted future locations serve as hints for the map decoder to focus on the area of perception features with high possibility. The history prior and future guidance provide a good initialization to track queries, which could be better aware of the perception features for accurate map instance localization.

\subsection{Semantic-Aware Query Generator}
\label{SAQG}
Recent approaches \cite{PivotNet, MapTR, MapTRv2, MapVR, MapTracker} treat map construction as a set prediction problem where each query is responsible for generating one map component. These learnable query-based decoding frameworks directly predict vectorized map elements by decoding queries from BEV features in parallel. However, these queries are typically initialized randomly and lack explicit alignment with the scene context, limiting their capacity to jointly encode semantic and geometric cues of map instances in complex environments. To overcome this limitation, we introduce a Semantic-Aware Query Generator that captures global semantic context to guide query generation more effectively.

We adopt a semantic-aware query generation strategy based on the Mask Transformer architecture introduced in Mask2Former \cite{Mask2Former}. Specifically, a set of learnable detection queries is initialized and progressively refined through $L$ layers of a Transformer decoder. At each decoding layer $l$, we apply mask-attention between the queries and the multi-scale BEV feature maps $F_{bev}$ based on the segmentation masks $M_{l-1}=\left\{ M_{q,l-1} \right\}^{N_q}_{q=1}$ produced in the $\left( l-1 \right)$-th layer. The corresponding semantic-aware (SA) detection queries $Q^{SA}_{l-1}=\left\{ Q^{SA}_{q,l-1} \right\}^{N_q}_{q=1}$ are then updated to capture both spatial and semantic context for subsequent decoding:

\begin{equation}
\hat{M}_{l-1} =\left\{ \begin{aligned}
0, &&  \text{if} \  M_{l-1}>\tau_L \\
-\infty  && \text{otherwise}
\end{aligned} \right.   
\end{equation}
\begin{equation}
Q_l=Q^{SA}_{l-1}W^Q,  K_l=F^{l}_{bev}W^K,   V_l=F^{l}_{bev}W^V 
\end{equation}
\begin{equation}
Q^{SA}_{l}=\text{softmax}\left( \hat{M}_{l-1}+Q_lK^T_l \right)V_l + Q^{SA}_{l-1}
\end{equation}
where $\tau_L$ is a mask threshold, and $W^Q$, $W^K$, $W^V$ are learnable weight matrices. After the last decoder layer, we apply dot product between the resulting semantic-aware queries $Q^{SA}_{L}$ and the BEV feature $F_{bev}$  along the channel axis to produce the instance-level segmentation masks $M_L$. These masks are subsequently used to update the instance-level History Maps.

\subsection{History Rasterized Map Memory}
\label{memory}
Similar to the Global Rasterized Map introduced in HRMapNet \cite{zhang2024enhancing}, we aim to create a History Rasterized Map Memory to store the historical instance-level segmentation masks. Unlike HRMapNet \cite{zhang2024enhancing}, which requires an additional rasterization function to convert predicted vectorized maps into rasterized form, our method directly generates rasterized map through the Semantic-Aware Query Generator. This eliminates the post-processing and preserves the end-to-end differentiability of the framework.

\noindent
\textbf{Initialization and Update.} At the beginning of frame $t$, we define the history rasterized map memory as a set of predicted instance segmentation masks for every instance $i$, $\mathcal{M}^{t}=\left\{{M^t_i} \in \mathbb{R}^{H\times W}\right\}^{N_{track}}_{i=1}$, where each $M^t_i$ is the BEV segmentation mask of the $i$-th previously tracked map instance, and $H$ and $W$ denote the spatial dimensions of the BEV feature grid.

To incorporate the rasterized map outputs at the current timestamp $t$, the memory is updated as follows: for predicted instance $j$, we determine whether it corresponds to a previously tracked instance. If it is matched to a track query with tracking index $i$, we retrieve its previous history map $M^{t-1}_i$ and update it using a temporal decay mechanism:
\begin{equation}
M^t_i = (1-\beta) \cdot M^{t-1}_i + \beta \cdot M^t_j \cdot S^t_j
\end{equation}
where $\beta$ is the decay factor to control the balance between past and new prediction, $S^t_j$ is the confidence score. If the predicted instance $j$ is unmatched and its detection confidence exceeds $\tau_{d}$, we initialize the map memory as:
\begin{equation}
M^t_{i} = M^t_j \cdot S^t_j
\end{equation}

\noindent
\textbf{Temporal Alignment.} When using history maps to guide the current frame’s perception, misaligned memory will lead to semantic noise or feature misguidance. To align history memory to the next frame for further updating, each history rasterized map $M^t_i \in \mathcal{M}^{t}$ is warped to frame $t+1$ using the ego-motion transformation $\text{Trans}^{t\rightarrow t+1}_{ego}$, 
\begin{equation}
\hat{M}^{t+1}_i=\text{Warp}\left( M^{t}_i, \text{Trans}^{t\rightarrow t+1}_{ego}\right)
\end{equation}
where Warp(·) denotes grid-based transformation over the BEV domain.

After warping, any track query at frame $Q^{t+1}_{track}$ does not meet the tracking threshold 
$\tau_t$ is considered lost. Such instances are pruned from the memory to prevent accumulation of invalid tracks. This dynamic memory update maintains a clean and relevant set of tracked instances, balancing temporal persistence with adaptability to changing scenes.

\subsection{History-Map Guidance Module}
\label{HMG}
To enhance the perceptual consistency of tracked instances, we introduce the History-Map Guidance (HMG) Module, which leverages fine-grained historical instance maps to refine the propagated track queries at each frame. Unlike previous works \cite{StreamMapNet, MapTracker} that rely on implicit feature propagation, our design enables explicit memory-driven refinement by sampling semantically relevant features using spatial cues derived from historical masks.

Specifically, for each active track query $Q^{track}_i$ at time step $t$, we access its corresponding historical rasterized mask $M^t_i$. To ensure reliable guidance, we compute a valid pixel mask $\mathcal{P}^{val}_i$ to filter regions in history map via memory threshold $\theta$, preserving only high-confidence spatial areas:
\begin{equation}
\mathcal{P}^{val}_i =\left\{ \begin{aligned}
1, &&  \text{if} \  M^t_i>\theta \\
0  && \text{otherwise}
\end{aligned} \right.   
\end{equation}

We then use $\mathcal{P}^{val}_i$ to sample features from the BEV features, capturing complementary geometric and semantic cues. To enhance spatial awareness in the BEV domain, we enrich BEV features with a position embedding $\text{PE}_{bev} \in\mathbb{R}^{H \times W \times C}$:
\begin{equation}
F^{sampled}_{bev} = \mathcal{P}^{val}_i \cdot \left(F_{bev} + \text{PE}_{bev}\right)
\end{equation}

In addition to these positional cues, we embed a semantic class embedding $\text{CE}_i \in \mathbb{R}^C$ for each track query, which encodes its semantic category to enhance the semantic representation. This class embedding is added to $Q^{track}_i$ to provide category priors.

Finally, the track query interacts with the sampled BEV features via a cross-attention layer, yielding a temporally aligned track query for the current frame:
\begin{equation}
\hat{Q}^{track}_i = \text{CrossAttn}\left(Q^{track}_i+\text{CE}_i,F^{sampled}_{bev}\right)
\end{equation}

By explicitly integrating historical spatial and semantic priors, the History-Map Guidance Module provides reliable and fine-grained query refinement across time, significantly enhancing both temporal and spatial consistency in online HD map construction.

\subsection{Short-Term Future Guidance Module}
\label{STFG}
In current frameworks like MapTracker \cite{MapTracker}, the propagation of tracked queries is often reactive, i.e., relying only on previous-frame information and cross-attention with the current BEV features. This can lead to unstable predictions under rapid scene changes, occlusions, or sensor noise. To address these challenges, we propose a Short-Term Future Guidance (STFG) module that explicitly predicts and incorporates short-term future locations of map instances. By providing the model with explicit motion priors, the STFG module enhances the temporal stability of track queries, enabling more reliable and coherent map instance detection across consecutive frames.

\noindent
\textbf{Trajectory Prediction.} Inspired by VIP3D \cite{VIP3D}, we extend the current query-based detection and tracking framework to a query-based prediction, for predicting the short-term future of tracked map instances. Specifically, for each tracked map query, we maintain a history of decoded polylines over the past $n$ frames: $\left\{ P^{t-n+1}_i, P^{t-n+2}_i, \cdots , P^{t}_i\right\}$ where $P^{t'}_i \in \mathbb{R}^{N_p \times 2}$. These sequential polylines are stacked into a temporal sequence and fed into a lightweight MLP head to capture motion patterns and geometric evolution. The head predicts a point-wise offset  $\Delta P^{t \rightarrow t+1}_i \in \mathbb{R}^{N_p \times 2}$ representing its immediate motion toward the next frame. The predicted offsets are added to the current coordinates of each instance to obtain their absolute predicted locations. The future polyline is estimated as:
\begin{equation}
\hat{P}^{t+1}_i = P^{t}_i + \Delta P^{t \rightarrow t+1}_i
\end{equation}
By modeling temporal dynamics explicitly, the STFG module produces structured future predictions that maintain geometric consistency and enhance alignment across frames.

\noindent
\textbf{Fusion with Track Queries.}
Once the short-term future polyline $\hat{P}^{t+1}_i \in \mathbb{R}^{N_p \times 2}$ of $i$-th map instance is predicted, we encode each predicted point coordinate  to a high-dimensional space using a learnable positional embedding function $\phi$:
\begin{equation}
\text{PE}^{future}_i = \frac{1}{N_p}\sum_{1}^{N_p}\phi \left(\hat{P}^{t+1}_{i,k} \right), k=1,\cdots N_P
\end{equation}
here, we aggregate the point-wise embeddings to form a compact representation of the entire polyline and adopt mean pooling for simplicity and efficiency. This global future embedding $\text{PE}^{future}_i \in \mathbb{R}^C$ captures the predicted spatial  distribution and overall structure of the instance in the next frame. It acts as a temporal positional prior for guiding the query update.

To inject the predicted short-term future into the next frame’s query propagation, we fuse the current track query embedding $Q^t_{track}$ with its corresponding future embedding $\text{PE}^{future}_i$. This operation yields the updated track query $Q^{t+1}_{track}$  to be used in the Transformer decoder in timestamp $t+1$, which represents as follows:
\begin{equation}
Q^{t+1}_{track} =\text{Linear}\left( \left[ Q^t_{track},  \text{PE}^{future}_i\right]\right)
\end{equation}
where $\left[ \cdot\right]$ denotes concatenation. 

With this, we then enable the track query to carry both semantic context from previous frames and spatial priors derived from predicted motion. As a result, the decoder in the next frame is better guided to attend to spatial regions that are both semantically relevant and temporally consistent, avoiding implausible detection and inaccurate global map construction. 

\subsection{Training Loss}
Our framework builds on MapTracker \cite{MapTracker}, which serves as the primary baseline. We
keep BEV loss ($\mathcal{L}_{BEV}$) and VEC loss ($\mathcal{L}_{track}$) functions consistent with MapTracker. In addition, similar to Mask2Former \cite{Mask2Former}, we supervise the BEV segmentation masks produced by the Semantic-Aware Query Generator using the binary cross-entropy mask loss and the dice loss \cite{VNet} with $\lambda_{dice} = 2$ and $\lambda_{bce} = 1$.
\begin{equation}
\mathcal{L}_{seg}=\lambda_{dice}\cdot  \mathcal{L}_{dice}+\lambda_{bce}\cdot  \mathcal{L}_{bce}
\end{equation}
\label{loss}
To train the Short-Term Future Guidance Module, we introduce a trajectory prediction loss that aligns predicted future polylines $\hat{P}^{t+1}$ with ground-truth locations $P^{t+1}_{gt}$: 
\begin{equation}
\mathcal{L}_{pred}= \text{CD}\left( \hat{P}^{t+1}, P^{t+1}_{gt}\right)
\end{equation}
\label{loss}
where CD is Chamfer Distance.

Finally, following MapTRv2 \cite{MapTRv2}, we adopt an auxiliary depth prediction loss $\mathcal{L}_{depth}$ that improves 3D spatial reasoning in the image backbone. The overall loss is defined as the weighted sum of the above losses:
\begin{equation}
\mathcal{L}_{total}=\mathcal{L}_{BEV}+\mathcal{L}_{track}+\mathcal{L}_{seg}+\mathcal{L}_{pred}+\mathcal{L}_{depth}
\end{equation}
\label{loss}
\vspace{-2mm}
\begin{table*}[]
\centering
\begin{tabular}{l|c|ccc|cc|c}
\hline
Method        & Epoch & $AP_{ped}$             & $AP_{divider}$     & $AP_{boundary}$              & $m$AP  $\uparrow$         & C-$m$AP $\uparrow$        & FPS  \\ \hline
PivotNet \cite{PivotNet}     & 24    & 56.2          & 56.5 & 60.1          & 57.6          & -             & -    \\
MapTRv2   \cite{MapTRv2}    & 24    & 59.8          & 62.4 & 62.4          & 61.5          & 41.7          & 14.1 \\
HRMapNet  \cite{HRMapNet}    & 24    & 65.8          & 67.4 & 68.5          & 67.3          & 49.2          & 10.3 \\
StreamMapNet \cite{StreamMapNet} & 24    & 61.9          & 66.3 & 62.1          & 63.4          & 38.4          & 13.1 \\
MGMap  \cite{MGMap}    & 24    & 61.8          & 65.0 & 67.5          & 64.8          & 43.5          & 13.4  \\
Mask2Map  \cite{Mask2Map}    & 24    & 70.6          & 71.3 & \textbf{72.9}          & 71.6          & 55.8          & 9.2  \\
\textbf{PredMapNet (Ours)} & 24    &       \textbf{74.1}       &   \textbf{72.8}   &   72.6            &     \textbf{73.2}          &   \textbf{64.3}           &    10.1  \\ \hline
HRMapNet  \cite{HRMapNet}    & 110   & 72.0          & 72.9 & 75.8          & 73.6          & 61.4          & 10.3 \\
MapTRv2  \cite{MapTRv2}     & 110   & 69.3          & 68.5 & 70.3          & 69.5          & 50.5          & 14.1 \\
Mask2Map \cite{Mask2Map}     & 110   & 73.6          & 73.1 & 77.3          & 74.6          & 60.3          & 9.2  \\
MapTracker \cite{MapTracker}    & 72    & \textbf{80.0} & 74.1 & 74.1          & 76.1          & 69.1          & 10.9 \\
\textbf{PredMapNet (Ours)} & 72    & 75.2          & \textbf{76.5} & \textbf{79.0} & \textbf{76.9} & \textbf{69.7} & 10.1 \\ \hline
\end{tabular}
\caption{Comparison with SOTA methods on the nuScenes validation set (old split \cite{MapTRv2}).}
\label{tab1}
\end{table*}

\begin{table*}[]
\centering
\begin{tabular}{l|c|ccc|cc}
\hline
Method        & Epoch & $AP_{ped}$  & $AP_{divider}$     & $AP_{boundary}$              & $m$AP  $\uparrow$         & C-$m$AP $\uparrow$          \\ \hline
MapTRv2 \cite{MapTRv2}      & 24    & 62.9          & 72.1          & 67.1          & 67.4          & -             \\
HRMapNet \cite{HRMapNet}     & 30    & 65.1          & 71.4          & 68.6          & 68.3          & -             \\
StreamMapNet \cite{StreamMapNet} & 72    & 70.5          & 74.2 & 66.1          & 70.3          & -             \\
Mask2Map \cite{Mask2Map}     & 24    & 68.1          & 72.7          & 73.7          & 71.5          & -   \\
MGMapNet \cite{yang2024mgmapnet} & 24    & 71.3          & 76.0 & 73.1          & 73.6          & -             
\\ \hline
MapTracker  \cite{MapTracker}  & 35    & 76.9 & 79.9          & 73.6          & 76.8          & 68.3          \\
\textbf{PredMapNet (Ours)} & 35    & \textbf{77.2}          & \textbf{80.3} & \textbf{74.5} & \textbf{77.3} & \textbf{69.1} \\ \hline
\end{tabular}
\caption{Comparison with SOTA methods on the Argoverse2 validation set.}
\label{tab2}
\end{table*}

\section{Experiments}
In this section, we present the experimental results of PredMapNet. We first introduce the implementation details of PredMapNet, and then report results and compare with SOTA methods using two widely-used autonomous driving datasets: nuScenes \cite{NuScenes}, Argoverse2 \cite{Argoverse}. All ablation studies are based on nuScenes \cite{NuScenes} dataset. 
\subsection{Implementation Details}
We implement our framework on top of MapTracker \cite{MapTracker} codebase. ResNet50 \cite{ResNet} is used as the image backbone. Training is conducted on nuScenes \cite{NuScenes} and Argoverse2 \cite{Argoverse} using 4 NVIDIA RTX A100 GPUs. On nuScenes, the model is optimized for 72 epochs in three stages (18, 6, and 48 epochs), while on Argoverse2, training is performed for 35 epochs (12, 3, and 20 epochs for the three stages). Key hyperparameters are set as follows: $N_q = 100$, $N_p = 20$, $C = 512$, $\tau_{d} = 0.4$, $\tau_{t} = 0.5$, $\tau_L = 0.5$, $\beta = 0.9$, $\theta = 0.5$.

\subsection{Datasets and Evaluation Metrics}
\noindent
\textbf{Datasets.} We evaluate PredMapNet on two real-world driving datasets: nuScenes and Argoverse2. The nuScenes \cite{NuScenes} dataset contains 1000 sequences of recordings collected by autonomous driving cars. Each episode is annotated at 2Hz and contains 6 camera images and LiDAR sweeps. The Argoverse2 dataset \cite{Argoverse} includes 1,000 logs. Each log provides 15s of 20Hz RGB images from 7 cameras and a log-level 3D vectorized map. Considering the overlap issue \cite{StreamMapNet} found in the original datasets, experiments were conducted on both the old \cite{MapTRv2} and new \cite{StreamMapNet} dataset splits for comprehensive evaluation.

\noindent
\textbf{Evaluation Metrics.} Following previous work \cite{MapTR,MapTRv2,MapTracker,StreamMapNet, Mask2Map},  We adopt two evaluation metrics: Average Precision (AP) based on Chamfer distance proposed in \cite{HDMapNet} and AP based on rasterization proposed in \cite{MapVR}. Evaluation thresholds are set at 0.5m, 1.0m, and 1.5m for mean AP ($m$AP). For rasterization-based mean AP ($m\text{AP}^{\dagger}$), we measure intersection over union for each map instance, with thresholds set \{0.50, 0.55, ..., 0.75\} for pedestrian crossings and \{0.25, 0.30, ..., 0.50\} for line-shaped elements. In addition, we employ consistency-aware metric (C-$m$AP) \cite{MapTracker} to penalize temporally inconsistent reconstructions.

\begin{table}[]
\centering
\resizebox{0.48\textwidth}{!}{\begin{tabular}{l|c|ccc|c}
\hline
Method   & Epoch & $AP_{ped}^{\dagger}$     & $AP_{divider}^{\dagger}$    & $AP_{boundary}^{\dagger}$    & $m\text{AP}^{\dagger}$    \\ \hline
MapTR  \cite{MapTR}  & 24    & 32.4 & 23.5 & 17.1 & 24.3 \\
$\text{MapTRv2}^{\star}$  \cite{MapTRv2} & 24    & 49.9 & 34.7 & 25.7 & 36.7 \\
Mask2Map \cite{Mask2Map} & 24    & 62.9 & 52.3 & 48.9 & 54.7 \\ \hline
\textbf{PredMapNet (Ours)}     & 24    &   69.3   &   60.1   &   63.5   & \textbf{64.3} \\ \hline
\end{tabular}}
\caption{Comparison of SOTA methods on nuScenes validation set (old split \cite{MapTRv2}) with rasterization-based metric. The "$\star$" indicates results reproduced using public codes.}
\label{tabras}
\end{table}

\subsection{Comparison with the State-of-the-art Methods}
\label{compa}
\subsubsection{\textbf{Results on nuScenes old split}}

Table \ref{tab1} reports the comparison with state-of-the-art methods on the nuScenes \cite{NuScenes} validation set (old split \cite{MapTRv2}). In the 24-epoch experiments, PredMapNet achieves 73.2 $m$AP, outperforming Mask2Map \cite{Mask2Map} by +1.6 $m$AP. In terms of temporal consistency, PredMapNet reaches 64.3 C-$m$AP, which represents a substantial gain of +8.5 over Mask2Map \cite{Mask2Map}. With extended training of 72 epochs, our model further improves to 76.9 $m$AP and 69.7 C-$m$AP, showing new state-of-the-art results. Compared to single-frame frameworks such as Mask2Map \cite{Mask2Map} and MGMap \cite{MGMap}, which lack explicit temporal modeling, PredMapNet consistently delivers superior accuracy and stability across all evaluation metrics. The improvements are especially notable in  C-$m$AP, indicating that incorporating both historical and short-term future guidance effectively enhances temporal coherence. When compared against MapTracker \cite{MapTracker}, our framework achieves higher accuracy while offering comparable efficiency.  During inference, PredMapNet runs at 10.1 FPS, slightly slower than MapTracker (10.9 FPS) due to the added feature sampling and short future forecasting. This overhead, however, can be alleviated with parallel acceleration, ensuring practicality for real-time deployment. In summary, PredMapNet not only surpasses previous SOTA methods in accuracy and temporal stability but also maintains computational efficiency, validating its effectiveness for practical HD map construction in autonomous driving systems. Table \ref{tabras} evaluates the performance of PredMapNet based on a rasterization-based metric. Notably, our method achieves a remarkable performance gain of 27.6 $m\text{AP}^{\dagger}$ over MapTRv2 \cite{MapTRv2} and outperforms Mask2Map \cite{Mask2Map} by 9.6 $m\text{AP}^{\dagger}$.

\begin{figure*}[t]
\centering
\includegraphics[width=\textwidth]{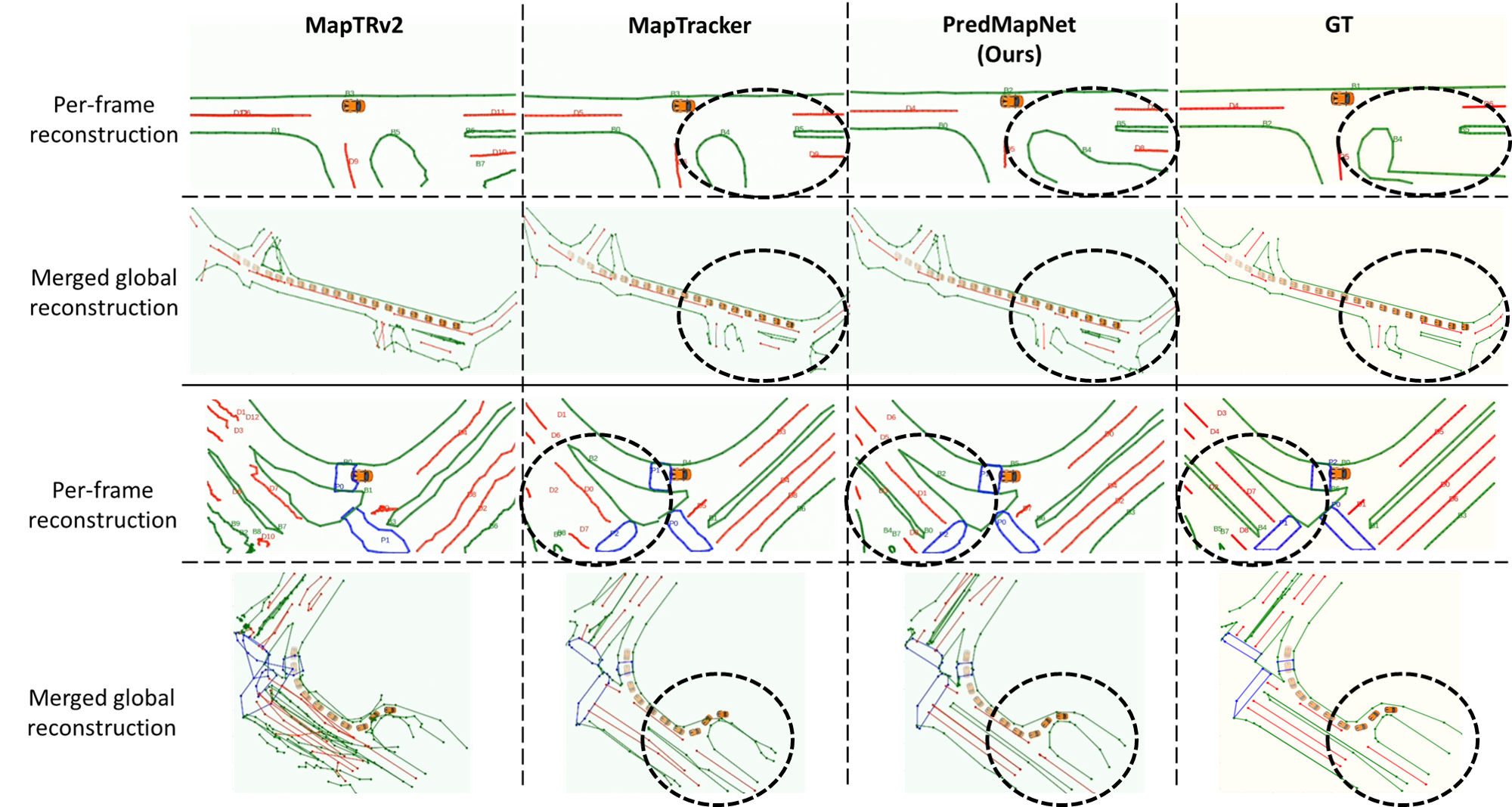}
\caption{Qualitative visualization on nuScenes val set.}
\label{vis}
\end{figure*}

\subsubsection{\textbf{Results on Argoverse2}}
The results on the Argoverse2 dataset \cite{Argoverse}, as shown in Table \ref{tab2}, further validate the effectiveness of PredMapNet. PredMapNet achieves 77.3 $m$AP and 69.1 C-$m$AP after 35 training epochs, outperforming existing methods across all metrics. Compared to StreamMapNet \cite{StreamMapNet} and Mask2Map \cite{Mask2Map}, our model delivers large improvements of +7.0 and +5.8 $m$AP, respectively. Relative to MapTracker \cite{MapTracker}, PredMapNet further provides consistent gains of +0.5 $m$AP and +0.8 C-$m$AP, confirming its ability to enhance both accuracy and temporal consistency across different scenarios.

\begin{table}[]
\centering
\begin{tabular}{c|c|c|c}
\hline
Dataset                     & Method                          & $m$AP  $\uparrow$                       & C-$m$AP  $\uparrow$ \\ \hline
\multirow{3}{*}{nuScenes}   & \multicolumn{1}{c|}{StreamMapNet \cite{StreamMapNet}}  & \multicolumn{1}{c|}{33.5} & 22.2  \\ 
                            & \multicolumn{1}{c|}{MapTracker \cite{MapTracker}} & \multicolumn{1}{c|}{40.3} & 32.5  \\
                            & \multicolumn{1}{c|}{\textbf{PredMapNet (Ours)}}       & \multicolumn{1}{c|}{\textbf{42.1}} & \textbf{33.7}  \\ \hline
\multirow{3}{*}{Argoverse2} & \multicolumn{1}{c|}{StreamMapNet \cite{StreamMapNet}}  & \multicolumn{1}{c|}{64.4} & 54.4  \\ 
                            & \multicolumn{1}{c|}{MapTracker \cite{MapTracker}} & \multicolumn{1}{c|}{70.3} & 61.3  \\
                            & \multicolumn{1}{c|}{\textbf{PredMapNet (Ours)}}       & \multicolumn{1}{c|}{\textbf{71.2}} & \textbf{62.3}  \\ \hline
\end{tabular}
\caption{Comparisons on non-overlapping datasets.}
\label{tab3}
\end{table}

\subsubsection{\textbf{Results on non-overlapping splits.}}
The nuScenes and Argoverse2 datasets exhibit geographical overlaps \cite{lilja2024localization}. StreamMapNet \cite{StreamMapNet} proposes a non-overlapping dataset split for them. The experimental results are shown in Table \ref{tab3}. Note that the performance for nuScenes degrades for all three methods. MapTracker \cite{MapTracker} consistently outperforms StreamMapNet \cite{StreamMapNet} with significant margins. Our method further surpasses MapTracker \cite{MapTracker}, achieving improvements of +1.8 $m$AP, +1.2 C-$m$AP, and +0.9 $m$AP, +1.0 C-$m$AP on Argoverse2.

\begin{table}[]
\centering
\resizebox{0.48\textwidth}{!}{\begin{tabular}{l|ccc|c}
\hline
                            & $AP_{ped}$  & $AP_{divider}$     & $AP_{boundary}$              & $m$AP  $\uparrow$    \\ \hline
Baseline                    & 77.3 & 72.4 & 74.2 & 74.7 \\
+Semantic-Aware Query \ref{SAQG}       & 77.2 & 73.1 & 74.7 & 75.0 \\
+History-Map Guidance  \ref{HMG}     & 77.6 & 73.6 & 75.0 & 75.4 \\
+Short-Term Future Guidance \ref{STFG} & 78.1 & 74.7 & 76.1 & 76.3 \\
+Aux Depth Supervision      & 79.0 & 75.2 & 76.5 & 76.9 \\ \hline
\end{tabular}}
\caption{Ablation study of main components of PredMapNet}
\label{tab4}
\end{table}
\vspace{-2mm}

\subsection{Ablation Studies}
We conducted an ablation study to evaluate the contributions of the core ideas of PredMapNet. Training was conducted on the nuScenes training dataset (old split \cite{MapTRv2}) for 72 epochs. Evaluation was also performed on the old split validation set.

\noindent
\textbf{Contributions of Main Components.} Table \ref{tab4} demonstrates the impact of each component of
PredMapNet. We evaluated performance by adding each component one by one. The first row represents a baseline model using MapTracker \cite{MapTracker}, which achieves 74.7 $m$AP. Adding the Semantic-Aware Query Generator (SAQG) improves map consistency by providing context-aligned detection queries derived from BEV masks. With this module, we observe a +0.3 gain in $m$AP and notable improvement in $AP_{divider}$  (72.4 → 73.1). This confirms the benefit of replacing randomly initialized queries with segmentation-guided semantic queries. Integrating the History-Map Guidance Module (HMG) yields a further +0.4 $m$AP boost. By leveraging temporally aligned historical maps, HMG refines track queries with fine-grained spatial priors, especially enhancing boundary and divider continuity. Furthermore, after incorporating Short-Term Future Guidance (STFG) module,  $m$AP raises to 76.3. STFG explicitly forecasts short-term motion and fuses it with current track queries, providing strong temporal priors. The +0.9 $m$AP improvement validates that future reasoning complements history-based priors and reduces implausible predictions. Lastly, applying an auxiliary depth supervision helps the backbone better encode 3D geometry, resulting in a higher performance of 76.9 $m$AP. This confirms that improving the 3D spatial understanding further benefits vectorized map construction.

\subsection{Qualitative Results}
Fig. \ref{vis} presents a qualitative comparison between PredMapNet and two state-of-the-art models, MapTRv2 \cite{MapTRv2} and MapTracker \cite{MapTracker}, on challenging scenes from the nuScenes validation set. For better visualization, we adopt the integration function from \cite{MapTracker} to accumulate per-frame predictions into a single global HD vectorized map. Ground-truth (GT) annotations are provided for reference. Across theses sample scenes, our approach exhibits stronger perceptual consistency and temporal alignment, especially in complex road geometries. The highlighted circle regions show key improvements: 1) our method more accurately captures the road boundaries and completes divider with improved geometric continuity. 2) PredMapNet produces smoother lane boundaries across a long-range stretch, where MapTracker introduces discontinuities and MapTRv2 yields noisy overlaps. These results validate the effectiveness of our proposed sub-modules in producing high-quality vectorized maps that are both spatially precise and temporally stable.
\section{Conclusion}
In this work, we present a novel end-to-end framework for consistent online vectorized HD map construction that integrates both historical and future reasoning. To address limitations in current query-based decoding pipelines, we propose three key modules: a Semantic-Aware Query Generator that enhances detection with global semantic cues, a History-Map Guidance Module that leverages fine-grained instance-level history map for spatial refinement, and a Short-Term Future Guidance Module that explicitly forecasts map instance trajectories to improve temporal continuity. By combining historical information with predictive guidance, our framework enables accurate and stable map instance localization and tracking across frames. Extensive experiments on nuScenes and Argoverse2 benchmarks demonstrate that our method consistently outperforms existing state-of-the-art approaches in both vectorized and rasterized map evaluation metrics, while maintaining practical inference efficiency. Our results validate the effectiveness of temporal priors in online mapping and provide a robust foundation for future research in global map construction for autonomous driving systems.
{
    \small
    \bibliographystyle{ieeenat_fullname}
    \bibliography{ref}
}

\end{document}



\appendix

\section{Appendix}

\subsection{Additional Results on nuScenes old split}

\subsubsection{The ablation study on history length of STFG} To analyze the impact of history length on the performance of our Short-Term Future Guidance (STFG) module, we conduct an ablation study by varying the number of historical frames used as input. Specifically, we evaluate performance on the nuScenes validation set (old split \cite{MapTRv2}) under 24-epochs training setting.

As shown in Table \ref{stfg_ablation}, using only 2 history frames yields a relatively weaker performance (72.1 $m$AP), suggesting that limited temporal context may not provide sufficient motion priors for stable prediction. As the number of history frames increases, the performance consistently improves, which highlights the importance of leveraging richer motion context for forecasting future positions of map instances. Notably, the best performance is achieved with 4 history frames, beyond which the benefit slightly declines. We hypothesize this is due to noise accumulation from longer temporal sequences, and outdated motion patterns that could mislead short-term reasoning. Hence, the 4-frames history offers the best balance between historical context and robustness, and is adopted as our final setup.

This experiment validates the effectiveness of STFG in leveraging motion priors to enhance the temporal consistency of map construction. Meanwhile, it confirms that incorporating a moderate temporal window provides an optimal balance between stability and responsiveness in short-term future guidance.

\begin{table}[ht]
\centering
\begin{tabular}{c|ccccc}
\toprule
\textbf{History Frames} & 2 & 3 & \textbf{4} & 5 & 6 \\
\midrule
$m$AP $\uparrow$   & 72.1 & 72.8 & \textbf{73.2} & 73.0 & 72.9 \\
C-$m$AP $\uparrow$  & 63.2 & 63.7 & \textbf{64.3} & 64.2 & 64.0 \\
\bottomrule
\end{tabular}
\caption{Ablation study on the number of history frames used in the Short-Term Future Guidance (STFG) module. Experiments are done on nuScenes old split for 24-epochs training.}
\label{stfg_ablation}
\end{table}

\begin{figure*}[htb]
\centering
\includegraphics[width=\textwidth]{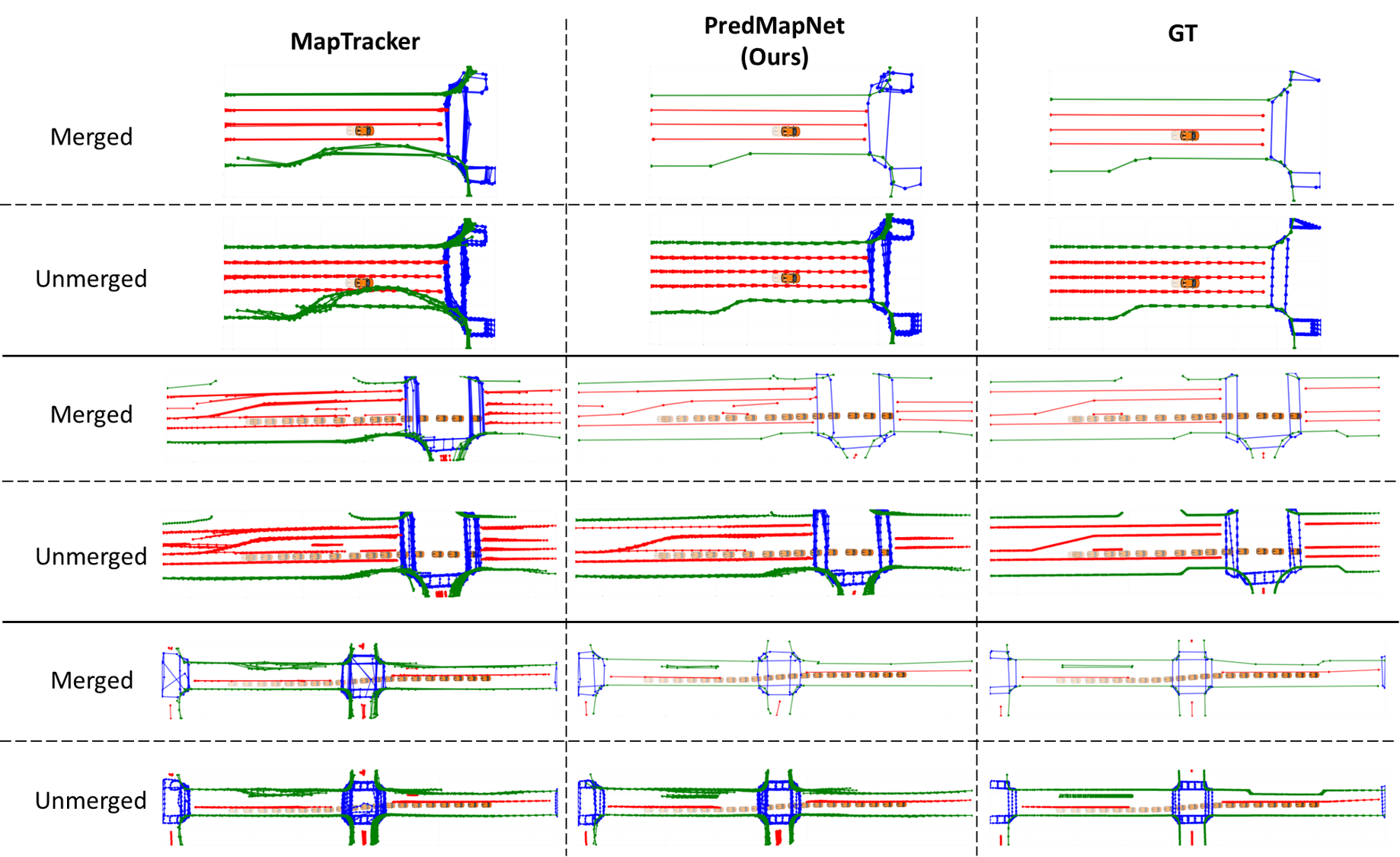}
\caption{Additional qualitative results on the Argoverse2 dataset.}
\label{visav}
\vspace{-1.5em}
\end{figure*}

\subsubsection{Qualitative Analysis}
Fig. \ref{visav} shows additional qualitative comparisons on Argoverse2 dataset. Our method demonstrates a marked improvement in producing smoother and temporally consistent vectorized map elements compared to MapTracker \cite{MapTracker}. Specifically, road boundaries (blue) and lane dividers (red) predicted by PredMapNet exhibit superior spatial continuity and alignment, particularly around complex junctions and curved segments. This is attributable to our History Map Guidance and Short-Term Future Guidance modules, which together inject both fine-grained historical context and motion-aware future priors into the detection pipeline for better query decoding.

Comparing each row of each example, our framework maintains connectivity across frames and generate more cleaner results. These qualitative gains demonstrate the effectiveness of leveraging both historical and future reasoning to refine the propagation and decoding of map queries. Overall, our design enhances the accuracy, and temporal stability of vector HD mapping in complex driving scenarios.


\subsection{Limitations and Potential Future Works}
While PredMapNet introduces novel components to enhance temporal consistency in online HD map construction, there remain some limitations that suggest directions for further improvement. Although the Short-Term Future Guidance (STFG) module brings motion priors to guide current-frame reasoning, our map decoder still operates in a single-frame decoding. While this design offers modularity and efficiency, it may still underexploit temporal continuity. Extending this design with temporally unified decoding could lead to stronger continuity and robustness, especially under challenging conditions such as occlusion and abrupt motion. Moreover, the current system lacks explicit mechanisms to recover from temporary object disappearance due to occlusion or sensor failure. Future work could incorporate occlusion-aware memory or recovery modules to allow query revival, improving stability in long-term deployments. While our model runs with reasonable efficiency, incorporating history map memory and future guidance modules introduces additional overhead. To ensure scalability in real-time autonomous driving applications, future work could explore lightweight architectural alternatives, such as query pruning or sparse attention mechanisms.

In conclusion, although our work is the first to integrate both history and short-term future reasoning into online HD map construction and achieves strong empirical results, future extensions could further improve long-range consistency, robustness in complex environments.

{    \small
    \bibliographystyle{ieeenat_fullname}
    \bibliography{ref}
}